# Belief Revision with Uncertain Inputs in the Possibilistic Setting

**Didier Dubois and Henri Prade**
Institut de Recherche en Informatique de Toulouse (IRIT) – CNRS
Université Paul Sabatier, 118 route de Narbonne
31062 Toulouse Cedex, France

**Abstract**

This paper discusses belief revision under uncertain inputs in the framework of possibility theory. Revision can be based on two possible definitions of the conditioning operation, one based on min operator which requires a purely ordinal scale only, and another based on product, for which a richer structure is needed, and which is a particular case of Dempster's rule of conditioning. Besides, revision under uncertain inputs can be understood in two different ways depending on whether the input is viewed, or not, as a constraint to enforce. Moreover, it is shown that M.A. Williams' transmutations, originally defined in the setting of Spohn's functions, can be captured in this framework, as well as Boutilier's natural revision.

## 1 INTRODUCTION

Belief revision in the sense of Gärdenfors (1988) is founded upon the existence of an epistemic entrenchment relation which rank-orders the formulas in a belief base to be revised. It has been pointed out that this epistemic entrenchment is nothing but a qualitative necessity measure and that faithful extensions of Gärdenfors' notions of expansion, contraction and revision can be defined in the framework of possibility theory and possibilistic logic; see (Dubois and Prade, 1992). In this setting, belief change can be either discussed in terms of a possibility distribution defined on the possible worlds (as in probability theory), or in terms of a possibilistic logic knowledge base (made of classical propositional formulas weighted by lower bounds of necessity measures) from which a possibility distribution can be defined.

Indeed, a layered consistent, propositional, belief base induces an ordering on the set of interpretations. Each layer may be interpreted as a level of certainty or as a specificity ranking in case of conditional knowledge (such as Z-ranking in the sense of Pearl (1990)). From such an ordering, encoded here as a possibility distribution $\pi$, a necessity measure N can be defined which enables us to recover the layer to which a formula belongs. More generally, an inference mechanism which propagates the certainty levels in agreement with the semantics provided through the underlying possibility distribution has been defined in the framework of possibilistic logic (Dubois et al., 1994). With this view in mind revising a layered knowledge base by the introduction of a new piece of information turns to be equivalent to some conditioning of the associated possibility distribution, while we remain in agreement with AGM postulates for revision (Gärdenfors, 1988). Moreover, any possibility distribution $\pi$ leads to a belief set (i.e., a deductively closed set of logical formula) $K = \{p \mid N([p]) > 0\}$, where $[p]$ denotes the set of models of proposition p, and the revised belief set by proposition q, $K^*_q$ is nothing but $\{p \mid N([p] \mid [q]) > 0\}$. In such a revision the input q is understood as a constraint expressing that q should be fully certain in the revised belief set, indeed $N([q] \mid [q]) = 1$. Following Boutilier (1993), we can consider that the input is only asserting that q should belong to the revised belief set, i.e., that in terms of a revised necessity measure $N^*_q$ we only require $N^*_q([q]) > 0$; this is called natural revision by Boutilier (who does not consider a revision process such that $N^*_q([q]) = 1$).

It is shown in the paper that natural revision is equivalent to several revisions under uncertain inputs in the possibilistic setting. Moreover, possibilistic revision under uncertain inputs, as discussed in Dubois and Prade (1993), is shown to exactly coincide with M.A. Williams' notion of adjustment. More generally, a unified view is provided, of various approaches developed by Boutilier (1993), Boutilier and Goldszmidt (1993), Williams (1994a, b; 1995), Goldszmidt and Pearl (1992), Darwiche and Pearl (1994). These works display two types of conditioning, that can be expressed in possibility theory, one (based on minimum operation) which is purely ordinal and is at work in Williams' adjustments for instance, and another (based on product, thus requiring a richer scale) alike Dempster's rule of conditioning, which is equivalent to Spohn (1988) conditioning through a rescaling and is used by Pearl and his coauthors. It is pointed out that belief revision with uncertain inputs can be expressed in terms of mixtures: convex sum/product mixtures in the probabilistic setting, and max-min or max-product mixtures in the possibilistic setting. Dubois et al. (1993) have introduced the qualitative counterpart of convex mixtures that also underlie qualitative decision theory (Dubois and Prade, 1995). We start by briefly recalling how belief change under uncertain inputs is dealt with in the probabilistic setting.

## 2 PROBABILISTIC SETTING

The Bayesian setting has been extended to the case of uncertain inputs under the form of a partition $\{A_1, A_2, ..., A_n\}$ of $\Omega$, and the probability attached to each $A_i$ is $\alpha_i$. The result of the revision is then (with $\Sigma\alpha_i = 1$ and $P(A_i) > 0$)

$$P(B \mid \{(A_i,\alpha_i)\}_{i=1,n}) = \Sigma_{i=1,n} \alpha_i P(B \mid A_i) \qquad (1)$$

It extends Jeffrey's rule:

$$P(B \mid (A,\alpha)) = \alpha P(B \mid A) + (1 - \alpha)P(B \mid \bar{A}) \qquad (2)$$

The revision rules (1) and (2) have been justified by P.M. Williams (1980) on the basis of minimizing the informational distance $I(P,P')$ under the constraints $P'(A_i) = \alpha_i$ for $i = 1,n$. (1) can also be justified at the formal level by the fact that the only way of combining the conditional probabilities $P(B \mid A_i)$ in an eventwise manner (i.e., using the same combination law for all events B) is to use a linear weighted combination such as (1) (Lehrer and Wagner, 1981). Note that the uncertain input is really viewed as a constraint that forces the probability measure to bear certain values on a partition of $\Omega$. The input assigns new probability values to some propositions $p_i$ with $[p_i] = A_i$ in a partition of $\Omega$. Jeffrey's rule ensures that probabilities do not change in relative value for situations within each partition element $A_i$.

In the above approach, the coefficient $\alpha_i$ is interpreted as the sure claim that the probability of input $A_i$ is $\alpha_i$; and leads to a correction of the prior probability. Especially it is a genuine revision process since the a priori probability and the input are at the same level, for instance they are both generic knowledge. However, consider the case where the input $\{(A_i,\alpha_i)\}_{i=1,n}$ means that an event has occurred that informs about the current situation, and $\alpha_i$ is the probability that this particular event is $A_i$. The input is really a piece of uncertain evidence and (1) is then the expected value of the conditional probability $P(B \mid A)$ where A is a random event whose realizations belong to the partition $\{A_1, ..., A_n\}$. This random event A is then a genuine unreliable observation, $\alpha_i$ being the probability that $A_i$ is *the* true input (and not the true probability of $A_i$). In other words, the coefficients represent a *complete* probability assignment on the family of subsets $2^\Omega$, and $\alpha_i = P'(\{A_i\})$ while in Jeffrey's rule, $\{(A_i,\alpha_i)\}_{i=1,n}$ is an *incompletely described* probability measure $P'$ on $\Omega$ and $\alpha_i = P'(A_i)$. It may look strange that the two views, corresponding to a revision and an uncertain focusing respectively, coincide in their implementation. However the requirement that $\{A_1, ..., A_n\}$ forms a partition does not look compulsory when the input is viewed as an uncertain (random) observation rather than a constraint on probability values (since it must just be a probability assignment on $2^\Omega$). The partition is compulsory only when the uncertain input is a constraint in order to ensure that the result obeys the constraint $P(A_i \mid \{(A_i,\alpha_i), i = 1,n\}) = \alpha_i$ (since $\Sigma_{i=1,n} \alpha_i = 1$). The uncertain input $(A,\alpha)$ interpreted as "event A has probably been observed (but maybe nothing has been observed)", corresponds to the probability assignment $\{(A,\alpha), (\Omega, 1 - \alpha)\}$ and not $\{(A,\alpha), (\bar{A}, 1 - \alpha)\}$ as for Jeffrey's rule. Indeed $P(\{A\}) = \alpha$ does not imply that $P(\{\bar{A}\}) = 1 - \alpha$.

## 3 POSSIBILISTIC SETTING

### 3.1 Basic Notions

The possibilistic approach consists in a slight improvement of the pure logical setting from the point of view of expressiveness. Instead of viewing a belief state as a flat set of mutually exclusive situations, one adds a complete partial ordering on top, according to which some situations are considered as more plausible than others. A cognitive state can then be modelled by a possibility distribution $\pi$, that is, a mapping from $\Omega$ to a totally ordered set V containing a greatest element (denoted 1) and a least element (denoted 0), typically the unit interval $V = [0,1]$. However any finite, or infinite and bounded, chain will do as well. The advantage of using the plausibility scale V is that it makes it easier to compare cognitive states. The idea of representing a cognitive state via a plausibility ordering on a set of situations ("possible worlds") is also pointed out by Grove (1986) and systematically used by Boutilier in his works (e.g., Boutilier, 1993).

Similarly to the probabilistic case, a possibility distribution generates a set function $\Pi$ called a possibility measure (Zadeh, 1978) defined by (for simplicity $V = [0,1]$)

$$\Pi(A) = \max_{\omega \in A} \pi(\omega) \qquad (3)$$

and satisfying $\Pi(A \cup B) = \max(\Pi(A),\Pi(B))$ as a basic axiom. The degree of certainty of A is measured by means of the dual necessity function $N(A) = 1 - \Pi(\bar{A})$.

The idea of revision is to get a consistent cognitive state even if the input information contradicts the a priori cognitive state. Revision in possibility theory is performed by means of a conditioning device similar to the probabilistic one, obeying an equation (à la Cox):

$$\forall B, \Pi(A \cap B) = \Pi(B \mid A) * \Pi(A). \qquad (4)$$

that is similar to Bayesian conditioning, together with $N(B \mid A) = 1 - \Pi(\bar{B} \mid A)$. Possible choices for $*$ are min and the product (the latter makes sense only in the numerical settings). However, for $* = $ min this equation may have more than one solution $\Pi(B \mid A)$. The least specific solution to (4) is chosen, (i.e., the solution with the greatest possibility degrees in agreement with the constraint (4)). The possibility distribution underlying the conditional possibility measure $\Pi(\cdot \mid A)$ is defined by

$$\begin{aligned}\pi(\omega \mid A) &= 1 \text{ if } \pi(\omega) = \Pi(A), \omega \in A,\\ \pi(\omega \mid A) &= 0 \text{ if } \omega \notin A,\\ &= \pi(\omega) \text{ if } \pi(\omega) < \Pi(A), \omega \in A.\end{aligned} \qquad (5)$$

Note that the existence of situations $\omega$ such that $\pi(\omega) = \Pi(A)$ is no longer guaranteed in the infinite case, and is generally added as an extra condition to the possibility distribution; it is referred to in the literature in terms of *well-ranked* orderings, and leads to adding specific postulates for revision (Williams, 1994a). Moreover, viewed as belief sets, i.e., deductively closed sets of logical formulae, based on a language whose interpretations form the set $\Omega$, any possibility distribution $\pi$ leads to a belief set $K = \{p,$ such that $N([p]) > 0\}$ (whose models are the core of $\pi$) and it is not difficult to check that the revised belief set $K^*_q$ using as input the formula q whose models form the set $[q] = A$ is $K^*_q = \{p,$ such that $N([p] \mid A) > 0\}$ (whose set of models is $\{\omega, \pi(\omega) = \prod(A)\}$ whenever $\prod(A) > 0$. This remark (also made by Williams, 1994b) points out that while the AGM theory revises belief sets, the possibilistic revision also revises the epistemic ordering of situations. Clearly iterated revision then becomes possible.

The above discussion of conditional possibility, using $* =$ min, makes sense in a purely qualitative setting. In a quantitative setting, $* =$ product may sound more reasonable and the expression corresponding to (4) is

$$\forall B, \prod(B \mid A) = \frac{\prod(A \cap B)}{\prod(A)} \quad (6)$$

provided that $\prod(A) \neq 0$. This is formally Dempster rule of conditioning, specialized to possibility measures, i.e., consonant plausibility measures in the sense of Shafer. The corresponding revised possibility distribution is

$$\pi(\omega \mid A) = \frac{\pi(\omega)}{\prod(A)}, \forall \omega \in A; \pi(\omega \mid A) = 0 \text{ otherwise.} \quad (7)$$

Note that for both (5) and (7), $N(A) = 1 \Rightarrow \pi(\cdot \mid A) = \pi$ (no revision takes place when the input information is already known with certainty). It can be shown (Dubois and Prade, 1992) that if $\pi^*_A$ denotes a possibility distribution obtained by revising $\pi$ with input A according to the AGM postulates, it makes sense to let $\pi^*_A = \pi(\cdot \mid A)$ since counterparts of the AGM postulates for revision hold as well with the two definitions. Note that if A and B are not disjoint, it can be verified that $\pi(\cdot \mid (\mid A) \mid B) = \pi(\cdot \mid A \cap B)$ when iterating the revisions. However if $A \cap B = \emptyset$ then $\pi(\cdot \mid (\mid A) \mid B)$ is undefined.

Moreover (5) embodies a principle of minimal change in the sense of the Hamming distance between $\pi$ and $\pi'$ defined by $H(\pi,\pi') = \sum_{\omega \in \Omega} |\pi(\omega) - \pi'(\omega)|$ ($\Omega$ finite). It makes sense for a finite totally ordered possibility scale V, just mapping the levels to integers. $H(\pi; \pi(\cdot \mid A))$ is minimal under the constraint $N(A \mid A) = 1$ as long as there is a single situation $\omega_A$ where $\pi(\omega_A) = \prod(A)$ (Dubois and Prade, 1992). When there is more than one situation $\omega$ where $\pi(\omega) = \prod(A)$, the principle of minimal change leads to as many revision functions by selecting one situation $\omega_A$ where $\pi(\omega_A) = \prod(A)$ and letting $\pi^*_A(\omega_A) = 1$ and $\pi^*_A(\omega_A) = \prod(A)$ for other most possible situations in A. In that case $\pi(\cdot \mid A)$ is the envelope of these minimal change revisions. A minimal change information-theoretic justification for the numerical, product-based conditioning rule exists in terms of relative levels of plausibility.

The contraction of a possibility distribution with respect to $A \subseteq \Omega$ corresponds to forgetting that A is true if A was known to be true. The result $\pi^-_A$ of the contraction must lead to a possibility measure $\prod^-_A$ such that $\prod^-_A(A) = \prod^-_A(\bar A) = 1$, i.e., complete ignorance about A. Intuitively if $\prod(A) = \prod(\bar A) = 1$ already, then we should have $\pi^-_A = \pi$. Besides if $\prod(A) = 1 > \prod(\bar A)$ then we should have $\pi^-_A(\omega) = 1$ for some $\omega$ in $\bar A$, and especially for those $\omega$ such that $\prod(\bar A) = \pi(\omega)$. It leads to (Dubois and Prade, 1992)

$$\pi^-_A(\omega) = 1 \text{ if } \pi(\omega) = \prod(\bar A), \omega \notin A$$
$$= \pi(\omega) \text{ otherwise.} \quad (8)$$

By construction, $\pi^-_A$ again corresponds to the idea of minimally changing $\pi$ so as to forget A, when there is a unique $\omega \in \bar A$ such that $1 > \prod(\bar A) = \pi(\omega)$. When there are several elements in $\{\omega \notin A, \pi(\omega) = \prod(\bar A)\}$, minimal change contractions correspond to letting $\pi^-_A(\omega) = 1$ for any selection of such situation, and $\pi^-_A$ corresponds to considering the envelope of the minimal change solutions. If $\prod(\bar A) = 0$, what is obtained is the fullmeet contraction (Gärdenfors, 1988). This contraction coincides exactly with a natural contraction in the sense of Boutilier and Goldzsmidt (1993).

### 3.2 Uncertain Inputs

An uncertain input information $(A,\alpha)$ is not understood in the same way whether it is a constraint or an unreliable input. In the first case, it forces the revised cognitive state to satisfy $N'(A) = \alpha$ (i.e., $\prod'(A) = 1$ and $\prod'(\bar A) = 1 - \alpha$) and the following belief change Jeffrey-like rule respects these constraints

$$\pi(\omega \mid (A,\alpha)) = \max(\pi(\omega \mid A), (1 - \alpha) * \pi(\omega \mid \bar A)) \quad (9)$$

where $* =$ min or product according to whether $\pi(\omega \mid A)$ is the ordinal or Bayesian-like revised possibility distribution. Note that when $\alpha = 1$, $\pi(\omega \mid (A,\alpha)) = \pi(\omega \mid A)$, but when $\alpha = 0$, we obtain a possibility distribution less specific than $\pi$, such that $N(A) = N(\bar A) = 0$.

When $\alpha > 0$ and $* =$ min, rule (9) exactly coincides with what Williams (1994b) calls an "adjustment" (see Subsection 3.4 below): the most plausible worlds in A become fully plausible, the most plausible situations in $\bar A$ are forced to level $1 - \alpha$ and all situations that were originally more plausible than $1 - \alpha$, if any are forced to level $1 - \alpha$ as well. This operation minimizes changes of the possibility levels of situations so as to accommodate the constraint $N'(A) = \alpha$. Williams (1994b) points out that for adjustments, if an event B is such that $N(B) > \max(N(A), N(\bar A), \alpha)$, then $N'(B) = N(B \mid (A,\alpha)) = N(B)$. In other words, firmly entrenched beliefs are left untouched.

Rule (9) can be extended to a set of input constraints $\prod(A_i) = \lambda_i$, $i = 1,n$, where $\{A_i, i = 1,n\}$ forms a partition of $\Omega$, such that $\max_{i=1,n} \lambda_i = 1$ (normalisation). It gives the following rule where $* =$ minimum or product whether $\pi(\omega \mid A_i)$ is ordinal or numerical:

$$\pi(\omega \mid \{(A_i,\lambda_i)\}) = \max_i \lambda_i * \pi(\omega \mid A_i). \quad (10)$$

In the second case $(A,\alpha)$ is viewed as an unreliable input, represented by the weighted *nested* pair of subsets $F = \{(A,1), (\Omega, 1 - \alpha)\}$ where the weights denote degrees of possibility. The revised cognitive state $\pi(\cdot \mid F)$ is defined by formal analogy with a probabilistic mixture as

$$\pi(\omega \mid F) = \max(\pi(\omega \mid A), \pi(\omega) * (1 - \alpha)). \quad (11)$$

Note the difference with (9): there is no conditioning on $\bar A$ ($\pi(\omega) = \pi(\omega \mid \Omega)$). However, contrary to (9), the equality $N(A \mid F) = \alpha$ is not warranted since $N(A \mid F) = N(A)$ whenever $N(A) > \alpha$. Lastly, $\pi(\omega \mid F) = \pi(\omega)$ if $\alpha = 0$ since then $F = \Omega$: no revision takes place. This behavior is very different from the case when the uncertain input is taken as a constraint.

Belief revision rule (9) has been proposed by Spohn (1988), using an ordinal conditional function $\kappa$ valued on the set $\mathbb{N}$ of natural integers such that $\forall A \subseteq \Omega$, $\kappa(A) = \min\{\kappa(\omega) \mid \omega \in A\}$ and $\kappa(\omega)$ can be viewed as a degree of impossibility of $\omega$. Letting $\prod_\kappa(A) = 1 - N_\kappa(\bar{A}) = 2^{-\kappa(A)}$, it is easy to check that $\pi_\kappa(\omega)$ is equal to $2^{-\kappa(\omega)}$, where $\pi_\kappa$ is the possibility distribution associated with $\prod_\kappa$. Spohn (1988) defines two conditioning concepts:

- $\forall \omega \in A$, $\kappa(\omega \mid A) = \kappa(\omega) - \kappa(A)$
- the (A,n)-conditionalization of $\kappa$ (conditioning by an uncertain input $\kappa'(A) = n$)
  $\kappa(\omega \mid (A,n)) = \kappa(\omega \mid A)$ if $\omega \in A$;
  $\kappa(\omega \mid (A,n)) = n + \kappa(\omega \mid \bar{A})$ if $\omega \in \bar{A}$.

whose possibilistic counterparts are equations (7) and (9) with $* =$ product:

$$\pi_\kappa(\omega \mid A) = \frac{\pi_\kappa(\omega)}{\prod_\kappa(A)} \text{ if } \omega \in A; \pi_\kappa(\omega \mid A) = 0 \text{ otherwise;}$$

$$\pi_\kappa(\omega \mid (A,n)) = \frac{\pi_\kappa(\omega)}{\prod_\kappa(A)} \text{ if } \omega \in A;$$

$$\pi_\kappa(\omega \mid (A,n)) = (1 - \alpha) \cdot \frac{\pi_\kappa(\omega)}{\prod_\kappa(A)} \text{ if } \omega \notin A.$$

With $\alpha = 1 - 2^{-n}$. The counterpart of (9) can be extended to an input ordinal conditional function $\kappa'$ defined on the partition $\{A_1, ..., A_n\}$: $\kappa(\omega \mid \kappa') = \kappa'(A_i) + \kappa(\omega \mid A_i)$, $\forall \omega \in A_i$, $i = 1,n$. This rule can be exactly mapped to the possibilistic belief change rule (10) where $* =$ product and $\lambda_i = 2^{-\kappa'(A_i)}$.

### 3.3 Boutilier's Natural Revision

Possibilistic conditioning is also coherent with Boutilier (1993)'s *natural revisions*. A natural revision by input A comes down to only assign to the most plausible situations in A a degree of possibility higher that other situations while retaining the same ordering of situations as before revision, including *for situations outside A*. This means that, after revision, some situations where A is not true may remain more plausible than situations where A is true. This feature does not fit the idea of revision via conditioning whereby, in the revised state, situations where the input information is false are deemed impossible (as is the case with probabilistic revision). The success postulate $N(A \mid A) = 1$ used in possibilistic revision is very strong, in accordance to the one in conditional probability ($P(A \mid A) = 1$). In Boutilier (1993)'s work an input A is not taken as definitely true but is only supposed to be accepted in the sense that in the revised state, only the condition $N^*_{nat A}(A) > 0$ holds for his so-called natural revision (where $N^*_{nat A}$ encodes the result of this revision). In the case of Boutilier (1993)'s natural revision, only the set of most plausible worlds in A are moved to become the overall most plausible states and is this sense it is still a minimal change revision. To describe this elementary change in possibilistic terms requires the use of non-normalized possibility distributions (so that $\prod(\bar{A}) > \prod(A)$, and $\prod^*_{nat A}(\bar{A}) = \prod(\bar{A})$ is always allowed, while $\prod^*_{nat A}(A) = 1 > \prod^*_{nat A}(\bar{A})$).

The enforced input $N(A) > 0$ can also be understood as "$N(A) = 1$ or $N(A) = \alpha_n$ or... or $N(A) = \alpha_2$" with scale $V$ be made of $n + 1$ levels $\lambda_1 = 1 > \lambda_2 > ... > \lambda_n > 0$, and $\alpha_i = 1 - \lambda_i$. It suggests that the natural revision can be expressed by a series of adjustments $\pi(\omega \mid (A, \alpha_i))$ considering their disjunction:

$$\pi^*_{nat A}(\omega) = \max_{i=2, n+1} \pi(\omega \mid (A, \alpha_i))$$
$$= \max(\pi(\omega \mid A), (1 - \alpha_2) * \pi(\omega \mid \bar{A})).$$

This rule coincides indeed with natural revision provided $* = \min$ and that no situation $\omega \in \bar{A}$ has a priori plausibility level $\pi(\omega) = \lambda_2$, so that when $\prod(A) < 1$, $\prod^*_{nat A}(\bar{A}) = \lambda_2 > \pi(\omega)$ for all $\omega$ such that $\pi(\omega) < 1$. Then natural revision comes down to raising the plausibility of the most plausible situations in A to 1 and forcing the most plausible situations in $\bar{A}$ down to $\lambda_2$ if they had plausibility 1 previously. If $\pi(\omega) = \lambda_2$ for situations $\omega \in \bar{A}$, it is always possible to let $\prod^*_{nat A}(\bar{A}) = \lambda'$ were $1 > \lambda' > \lambda_2$ if $V$ is infinite $(= [0,1])$. For $* =$ product this rule is found in Darwiche and Pearl (1994) under the name R-conditioning.

### 3.4 M.A. Williams' Approach

Williams (1994b) has defined a general form of belief change she calls "transmutations", in the setting of Spohn's functions. Given an uncertain input $(A,n)$ taken as a constraint and a Spohn function $\kappa$ describing the agent's a priori cognitive state, a transmutation of $\kappa$ by $(A,n)$ produces a Spohn function $\kappa'$ such that $\kappa'(\bar{A}) = n$ and $\kappa'(A) = 0$, i.e., the degree of acceptance of A is enforced to level n. Clearly, this notion makes sense in the possibilistic setting, where a transmutation of a cognitive state $\pi$ into $\pi'$ using input $N'(A) = \alpha$ corresponds to enforcing $N'(A) = \alpha$ and $N'(\bar{A}) = 0$. Williams (1994a) has introduced a more qualitative transmutation called an adjustment. An adjustment of $\kappa$ by $(A,n)$ is either a contraction $\kappa^-_A$ if $n = 0$ or another belief change operation, defined as follows:

$$\kappa^*_{(A,n)} = \kappa^-_A \text{ if } n = 0$$
$$\kappa^*_{(A,n)} = (\kappa^-_A)^x_{(A,n)} \text{ if } 0 < n < \kappa(\bar{A})$$
$$\kappa^*_{(A,n)} = \kappa^x_{(A,n)} \text{ otherwise}$$

where  $\kappa^-_A(\omega) = 0$ if $\omega \in \bar{A}$ and $\kappa(\omega) = \kappa(\bar{A})$
$\kappa^-_A(\omega) = \kappa(\omega)$ otherwise

$\kappa^x_{(A,n)}(\omega) = 0$ if $\omega \in A$ and $\kappa(\omega) = \kappa(A)$
$= \kappa(A)$ if either $\omega \in A$ and $\kappa(\omega) \neq \kappa(A)$
 or $\omega \in \bar{A}$ and $\kappa(\omega) > n$
$= n$ otherwise.

In order to clarify the meaning of this intricate definition of adjustment, let us map it to the possibilistic setting (with $\alpha = 1 - 2^{-n}$). It can be shown (see Appendix) that the result of this translation is precisely

$$\pi^*_{(A,\alpha)}(\omega) = \max(\pi(\omega \mid A), \min(1 - \alpha, \pi(\omega \mid \bar{A})))$$

where we recognize (9). This result leads us to simplifying Williams' adjustment as follows

$$\kappa^*_{(A,n)}(\omega) = \min(\kappa^*(\omega \mid A), \max(n, \kappa^*(\omega \mid \bar{A}))) \quad (12)$$

where $\kappa^*(\omega \mid A) = +\infty$ if $\omega \notin A$; $\kappa^*(\omega \mid A) = \kappa(\omega)$ if $\kappa(\omega) > \kappa(A)$; $\kappa^*(\omega \mid A) = 0$ if $\kappa(\omega) = \kappa(A)$. $\kappa^*(\omega \mid A)$ is the Spohnian version of the qualitative form of possibilistic conditioning.

## 4 SYNTACTIC REVISIONS OF POSSIBILISTIC BELIEF BASES

Revision tools developed at the semantic level can be expressed at the level of a knowledge base expressed in a possibilistic logic (Dubois, Lang and Prade, 1994).

### 4.1 Possibilistic Logic

Possibilistic logic syntax consists of sentences in the first order calculus to which are attached lower bounds on the degree of necessity (or possibility) of these sentences. In this section, degrees of uncertainty belong to a totally ordered set V with bottom 0 and top 1. Here we consider only the fragment of possibilistic logic with propositional sentences to which lower bounds of degrees of necessity are attached. If p is a propositional sentence, (p $\alpha$) is the syntactic counterpart of the semantic constraint $N([p]) \geq \alpha$.

A possibilistic belief base is a finite set $\mathcal{K} = \{(p_i \alpha_i), i=1,n\}$ of weighted (propositional) formulae that contain beliefs explicitly held by an agent. The weight indicates the agent's confidence in the corresponding formula. Note that any belief base $\mathcal{B}$ (i.e., set of propositional sentences) equipped with a complete partial ordering $\geq$ can be mapped to a possibilistic belief base, changing p and q in $\mathcal{B}$ into (p $\alpha$) and (q $\beta$) such that $\alpha \geq \beta$ if and only if p $\geq$ q. As already pointed out the unit interval could be changed into any bounded, totally ordered set; the possibility/necessity duality is then expressed by reversing the ordering.

Reasoning in possibilistic logic is done by means of an extension of the resolution principle to weighted clauses:

$$(c\ \alpha); (c'\ \beta) \vdash (\text{Res}(c,c')\ \min(\alpha,\beta))$$

where c and c' are propositional clauses, and Res(c,c') is their resolvent. For instance, $(\neg p \vee q\ \alpha); (p \vee r\ \beta) \vdash (q \vee r\ \min(\alpha,\beta))$. This inference rule presupposes that when in a possibilistic formula (p $\alpha$) p is not in a clausal form, it can be turned into a set $\{(c_i\ \alpha), i = 1,n\}$ of weighted clauses such that p is equivalent to $c_1 \wedge c_2 \wedge \ldots \wedge c_n$. This is justified by the semantics of propositional logic, and by the fact that $N(p) \geq \alpha$ is equivalent to $N(c_i) \geq \alpha$, for $i = 1,n$ (from now on, we write $N(p)$ instead of $N([p])$ for short). Inference from a possibilistic belief base is denoted $\mathcal{K} \vdash (p\ \alpha)$, and is short for $\mathcal{K} \cup \{(\neg p\ 1)\} \vdash (\bot\ \alpha)$ (refutation method). $\mathcal{K} \vdash (\bot\ \alpha)$ can be checked by means of repeated uses of the resolution principle until the empty clause is attained, with some positive weight. The degree of inconsistency inc($\mathcal{K}$) is then defined by $\max\{\alpha \mid \mathcal{K} \vdash (\bot\ \alpha)\}$.

The set of possible situations in which a possibilistic logic sentence (p $\alpha$) is true is a fuzzy set [p $\alpha$] on $\Omega$ defined by

$$\mu_{[p\ \alpha]}(\omega) = 1 \text{ if } \omega \in [p]$$
$$= 1 - \alpha \text{ otherwise,}$$

where [p] is the set of possible situations where p is true; $\mu_{[p\ \alpha]}$ is the least specific (i.e., the greatest) possibility distribution $\pi$ such that $N([p]) = \inf_{\omega \notin [p]} 1 - \pi(\omega) \geq \alpha$. A possibilistic belief base $\mathcal{K} = \{(p_i\ \alpha_i), i = 1,m\}$ (with the semantics $N([p_i]) \geq \alpha_i$) is represented by the possibility distribution

$$\pi(\omega) = \min_{i=1,m} \max(\mu_{[p_i]}(\omega), 1 - \alpha_i) \quad (13)$$

which extends $[\mathcal{K}] = [p_1] \cap [p_2] \cap \ldots \cap [p_n]$ from a set of sentences to a set of weighted sentences. $\pi$ is the least specific possibility distribution such that $\forall\ (p\ \alpha) \in \mathcal{K}$, $N(p) \geq \alpha$, where N is computed with $\pi$. Semantic entailment is defined in terms of specificity ordering ($\pi \leq \pi'$). Namely, $\mathcal{K} \vDash (p\ \alpha)$ if and only if $\pi \leq \max(\mu_{[p]}, 1 - \alpha)$. This notion of semantic entailment is exactly the one of Zadeh (1979). Note that the above framework can be equivalently expressed in terms of Spohnian functions. Instead of using weights in the unit interval, we can use integers from 0 to n where n is the number of layers in the ordered belief base. An obvious understanding of p being in layer i is that $\kappa(\neg p) \geq i$, which can be made equivalent to $N(p) \geq \alpha$, where $i = -\text{Log}_2(1 - \alpha)$. The most entrenched sentences are then in layer n (see for instance Williams, 1995). The minimally specific possibility distribution $\pi$ that is induced by an ordered belief base is then changed into a ranking of the possible words, namely, letting $\ell(p_i)$ the layer number of $p_i$

$$\kappa(\omega) = 0 \text{ if } \omega \text{ satisfies all } p_i\text{'s in } \mathcal{K}$$
$$= \max_{i: \omega \vDash \neg p_i} \ell(p_i) \text{ otherwise.}$$

This is called "minimal ranking function" by Pearl (1990) and it corresponds to the minimally specific possibility distribution $\pi$.

Possibilistic logic is sound and complete with respect to refutation based on resolution (Dubois et al., 1994). Namely, it can be checked that

$$\text{inc}(\mathcal{K}) = 1 - \gamma = 1 - \max_{\omega \in \Omega} \pi(\omega)$$
$$\mathcal{K} \vdash (p\ \alpha) \text{ if and only if } \mathcal{K} \vDash (p\ \alpha).$$

When inc($\mathcal{K}$) < 1, $\mathcal{K}$ is said to be partially inconsistent; when inc($\mathcal{K}$) = 0 it is completely inconsistent. Consistent possibilistic belief bases are such that $\pi(\omega) = 1$ for some $\omega \in \Omega$. Consistency of $\mathcal{K}$ is equivalent to the consistency of the classical knowledge base $\mathcal{K}^*$ obtained by removing the weights.

When $\mathcal{K}$ is consistent, we can define an ordered belief set generated by $\mathcal{K}$ as Cons($\mathcal{K}$) = {(p $\alpha$), $\mathcal{K} \vdash (p\ \alpha)$}. The set function N such that $N([p]) = \alpha$ for all (p $\alpha$) in Cons($\mathcal{K}$) (and 0 otherwise) is a necessity measure. The ordering $\geq_N$ generated on Cons($\mathcal{K}$) is (up to some limit conditions) an epistemic entrenchment ordering in the sense of Gärdenfors (1988) and an expectation ordering in the sense of Gärdenfors and Makinson (1994); see Dubois and Prade (1991). Note that the restriction to $\mathcal{K}$ of the ordering on Cons($\mathcal{K}$) may fail to respect the original ordering on $\mathcal{K}$. Indeed the restriction of a generated ordered belief set to a belief set $\mathcal{K}$ satisfies the so-called *EE-coherence* property:

For any (p $\alpha$) $\in \mathcal{K}$ and any subset $\mathcal{B} = \{(p_i\ \beta_i),$ $i = 1,m\}$ of $\mathcal{K}$,
if $\mathcal{B} \vdash$ (p $\alpha$) then it does not hold that $\alpha < \min_{i=1,m} \beta_i$.

The above condition is requested by Rescher (1976) and Rott (1991). The restriction of an epistemic entrenchment ordering to a belief base is called "ensconcement" by Williams (1994a). Starting with any ordering on a belief base, possibilistic inference restores the above condition on it.

When $\mathcal{K}$ is partially inconsistent, i.e., inc($\mathcal{K}$) > 0, non-trivial deductions can still be made from $\mathcal{K}$, namely all (p $\alpha$) such that $\mathcal{K} \vdash$ (p $\alpha$), and for which $\alpha >$ inc($\mathcal{K}$). Indeed (p $\alpha$) is then the consequence of a consistent subpart of $\mathcal{K}$. Non-trivial inference of p from $\mathcal{K}$ is denoted $\mathcal{K} \vdash_{pref}$ p. When $\mathcal{K}$ is partially (but not totally) inconsistent, the associated consistent ordered belief set is Cons$_{pref}(\mathcal{K}) = \{(p\ \alpha),\ \mathcal{K} \vdash_{pref} p\}$.

In that case $\max_{\omega \in \Omega} \pi(\omega) = 1 - $ inc($\mathcal{K}$) < 1 and $\forall p$, min(N(p),N($\neg$p)) = inc($\mathcal{K}$). Let $\tilde{\pi}$ be a possibility distribution on $\Omega$ defined by

$$\tilde{\pi}(\omega) = \pi(\omega) \text{ if } \pi(\omega) < 1 - \text{inc}(\mathcal{K})$$
$$= 1 \text{ otherwise}$$

then it is easy to verify that the necessity measures N and $\tilde{N}$ based on $\pi$ (defined by (13)) and $\tilde{\pi}$ respectively are related by the following relation

$$N(p) > N(\neg p) \Rightarrow \tilde{N}(p) = N(p) > N(\neg p)$$
$$= \text{inc}(\mathcal{K}) > \tilde{N}(\neg p) = 0.$$

In fact, we have that

$$\tilde{\pi}(\omega) = \min_{i:\alpha_i > \text{inc}(\mathcal{K})} \max(\mu_{[p_i]}(\omega), 1 - \alpha_i).$$

### 4.2 Revising Ordered Belief Bases

Belief change in possibilistic logic can then be envisaged as the syntactic counterpart of change for possibility distributions. Let $\mathcal{K}$ be a consistent set of possibilistic formulae. Expansion of $\mathcal{K}$ by p consists of forming $\mathcal{K} \cup \{(p\ 1)\}$, provided that inc($\mathcal{K} \cup \{(p\ 1)\}$) = 0. Clearly, the possibility distribution $\pi'$ that restricts the fuzzy set of situations that satisfy $\mathcal{K} \cup \{(p\ 1)\}$ is $\pi' = \min(\pi, \mu_{[p]})$.

Let us consider the case when $\mathcal{K}$ is consistent, but $\mathcal{K}' = \mathcal{K} \cup \{(p\ 1)\}$ is not, and let $\alpha = $ inc($\mathcal{K} \cup \{(p\ 1)\}$) > 0. The following identity is easy to prove (e.g., Dubois et al., 1994):

$$\mathcal{K} \cup \{(p\ 1)\} \vdash_{Pref} (q\ \beta) \text{ if and only if } N(q \mid p) > 0$$

where N(q | p) is the necessity measure induced from $\pi(\cdot \mid [p])$, i.e., the possibility distribution expressing the content of $\mathcal{K}$, revised with respect to the set of models of p. Indeed let $\pi'$ be the possibility distribution on $\Omega$ induced by $\mathcal{K}'$, then

$$\pi' = \min(\pi, \mu_{[p]})$$
$$0 < \max_{\omega \in \Omega} \pi'(\omega) = 1 - \alpha < 1$$

and the possibility distribution $\tilde{\pi}'$ induced from the consistent part of $\mathcal{K}'$ made of sentences whose weight is higher than $\alpha$, is defined as

$$\tilde{\pi}'(\omega) = \pi(\omega) \text{ if } \omega \in [p] \text{ and } \pi'(\omega) < 1 - \alpha$$
$$= 1 \text{ if } \omega \in [p] \text{ and } \pi(\omega) = 1 - \alpha$$
$$= \pi'(\omega) = 0 \text{ otherwise.}$$

Hence $\tilde{\pi}' = \pi(\cdot \mid [p])$, the result of revising $\pi$ by [p] using the ordinal conditioning method of Section 3.1.

The possibilistic revision rule based on ordinal conditioning can be expressed directly on the belief base $\mathcal{K}$ by the following method (Dubois and Prade, 1992), called "brutal theory base operator" by Williams(1994a) and then $\mathcal{K}*_p$ is obtained:

i)  adding p above the top layer of $\mathcal{K}$
ii) deleting all sentences whose level is below the inconsistency level $\alpha = $ inc($\mathcal{K} \cup \{(p\ 1)\}$).

This belief base revision is rather drastic since all sentences $(p_i\ \alpha_i)$ with weights $\alpha_i \leq \alpha$ are thrown away, and replaced by (p 1). However it is syntax-independent. Note that this revision method works even if the weights attached to formulae are not EE-coherent. Suppose (q $\beta$) $\in \mathcal{K}$ and $\mathcal{K} \vdash$ (q $\gamma$) with $\gamma > \beta$. It means that (q $\beta$) can be deleted from $\mathcal{K}$ without altering its fuzzy set of models. The revision being syntax-independent, the presence or the absence of (q $\beta$) in $\mathcal{K}$ will not affect the fuzzy set of models of $\mathcal{K}*_p$. Note that when $\Pi(p) > 0$, N(q | p) > 0 is equivalent to N($\neg p \vee q$) > N($\neg p \vee \neg q$), i.e., in terms of epistemic entrenchment (Gärdenfors, 1988), $\neg p \vee q$ is more entrenched than $\neg p \vee \neg q$, and corresponds to a characteristic condition for having q in the (ordered) belief set obtained by revising Cons($\mathcal{K}$) with respect to p, in Gärdenfors (1988). We just showed that this revision is easily implemented in the possibilistic belief base itself, without making the underlying ordered belief set explicit. This result goes against the often encountered claim that working with epistemic entrenchment orderings would be intractable. Note that the revision produces a new epistemic entrenchment ordering.

A more parsimonious revision scheme for possibilistic belief bases $\mathcal{K}$ receiving an input p is to consider all subsets of $\mathcal{K}$ that fail to infer ($\neg p\ \alpha$) for $\alpha > 0$. If $\mathcal{H}$ is such a subset, then the result of the revision could be $\mathcal{H} \cup \{(p,1)\}$. We may take advantage of the ordering in $\mathcal{K}$ to make the selection. Namely we may restrict ourselves to $\mathcal{H}$ such that $\forall$ (q $\alpha$) $\notin \mathcal{H}$, $\mathcal{H} \cup \{(q\ \alpha),$ (p 1)$\} \vdash (\bot\ \alpha)$ in the possibilistic logic sense, i.e., (q $\alpha$) is involved in the contradiction. This proposal, made independently in Dubois et al. (1992) corresponds to selecting a preferred subbase in the sense of Brewka (1989). This selection process leads to a unique solution if $\mathcal{K}$ is totally ordered. In case of ties, further refinement can be made using a lexicographic ordering of the weights of the sentences not in $\mathcal{H}$, as proposed in Dubois et al.

(1992). These revision processes, that also relate to Nebel (1992)'s syntax-based revision schemes, are systematically studied in Benferhat et al. (1993).

This kind of revision process cannot be expressed at the semantic level, where all sentences in the knowledge base $\mathcal{K} \cup \{(p\ 1)\}$ have been combined into a possibility distribution on $\Omega$, and revision is performed on the aggregated possibility distribution. Especially if $(q\ \beta) \in \mathcal{K}$ and $\beta < \text{inc}(\mathcal{K} \cup \{(p\ 1)\})$ then $\min(\pi, \mu_{[p]}) \leq \max(\mu_{[q]}, 1 - \beta)$, i.e., everything happens as if $(q\ \beta)$ had never been in $\mathcal{K}$. The alternative syntactic revision rule, explained above, breaks the minimal inconsistent subsets of $\mathcal{K} \cup \{(p\ 1)\}$ in a parsimonious way, enabling pieces of evidence like $(q\ \beta)$ to be spared when they are not involved in the inconsistency of $\mathcal{K} \cup \{(p\ 1)\}$.

**Example**: Consider the belief base $\mathcal{K} = \{(\neg p\ \alpha), (q\ \beta)\}$ with $\beta < \alpha$. Then
$\pi(\omega) = \min(\max(1 - \mu_{[p]}(\omega), 1 - \alpha), \max(\mu_{[q]}(\omega), 1 - \beta))$
$= 1$ if $\omega \models \neg p \wedge q$
$= 1 - \alpha$ if $\omega \models p$
$= 1 - \beta$ if $\omega \models \neg p \wedge \neg q$.

Revising by input p at the semantic level leads to consider

$$\pi'(\omega) = \min(\pi(\omega), \mu_{[p]}(\omega)) = \begin{cases} 1 - \alpha \text{ if } \omega \models p \\ 0 \text{ otherwise.} \end{cases}$$

Hence $\pi(\omega\ |\ [p]) = 1$ if $\omega \models p$ and $0$ otherwise. Hence $\mathcal{K}*_p = \{(p\ 1)\}$. Acting at the syntactic level, the preferred sub-base of $\{(p\ 1), (\neg p\ \alpha), (q\ \beta)\}$ that contains p is $\{(p\ 1), (q\ \beta)\}$. Note that although $\min(\pi, \mu_{[p]}) \leq \max(\mu_{[q]}, 1 - \beta)$ we no longer have $\pi(\omega\ |\ [p]) \leq \min(\mu_{[q]}(\omega), 1 - \beta)$, i.e., adding the low certainty formulas consistent with $\mathcal{K}*_p$ leads to a non-trivial expansion of $\pi(\cdot\ |\ [p])$. It points out the already mentioned weakness of the semantic views of revision, which is particularly true with numerical approaches: the representation of the cognitive state is lumped, i.e., the pieces of belief are no longer available and the semantic revision process cannot account for the structure of the cognitive state that is made explicit in the ordered belief base.

Possibilistic base revision can be extended to the case of uncertain inputs. This has been done by Williams (1995) for her so-called adjustments. We have pointed out that if the input information is of the form $(p\ \alpha)$, an adjustment is the possibilistic counterpart of Jeffrey's rule as given by equation (9) and $* = \min$ when the weight is positive. Williams (1995) gives a rather intricate recipe to achieve the adjustment on the belief base itself. Stated in terms of necessity measures, the adjustment comes down to computing, for all formulae q the degree

$$N(q\ |\ (p\ \alpha)) = \min\ (N(q\ |\ p), \max(\alpha, N(q\ |\ \neg p))).$$

Here we assume that the ordering in $\mathcal{K}$ is EE-coherent. For adjusting a belief base $\mathcal{K}$ to input $(p\ \alpha)$, it is enough to compute $\mathcal{K}*_p$ and $\mathcal{K}*_{\neg p}$ using the above brutal syntactic revision method. Then if a formula q appears in $\mathcal{K}*_p$ or in $\mathcal{K}*_{\neg p}$ let it be with respective weights $\beta^+$ and $\beta^-$, with the understanding that the weight is 0 if q does not appear in the corresponding belief base. The adjusted belief base $\mathcal{K}*_{(p\ \alpha)}$ is made of all these formulae, each of which is assigned the weight min $(\beta^+, \max(\alpha, \beta^-))$.

Possibility theory thus offers a framework where the connection between semantic and syntactic forms of belief change can be easily laid bare.

**Acknowledgements.** This work has been partially supported by the European ESPRIT Basic Research Action No. 6156 entitled "Defeasible Reasoning and Uncertainty Management Systems (DRUMS-II)".

Belief Revision in the Possibilistic Setting 243

## APPENDIX: Mapping adjustment of kappa functions to the possibilistic setting

An adjustment of $\kappa$ by $(A,n)$ is either a contraction $\kappa^-_A$ if $n = 0$ or another belief change operation, defined by

$$\kappa^*_{(A,n)} = \kappa^-_A \text{ if } n = 0$$
$$= (\kappa^-_A)^x_{(A,n)} \text{ if } 0 < n < \kappa(\bar{A})$$
$$= \kappa^x_{(A,n)} \text{ otherwise}$$

where $\kappa^-_A(\omega) = 0$ if $\omega \in \bar{A}$ and $\kappa(\omega) = \kappa(\bar{A})$
$= \kappa(\omega)$ otherwise

$\kappa^x_{(A,n)}(\omega) = 0$ if $\omega \in A$ and $\kappa(\omega) = \kappa(A)$
$= \kappa(A)$ if either $\omega \in A$ and $\kappa(\omega) \neq \kappa(A)$
  or $\omega \in \bar{A}$ and $\kappa(\omega) > n$
$= n$ otherwise.

It is obvious that $\kappa^-_A(\omega)$ becomes $\pi^-_A(\omega)$ using the mapping $\pi(\omega) = 2^{-\kappa(A)}$, and considering the definition of a contraction in the possibilistic setting. Hence Williams contraction exactly corresponds to possibilistic contraction. Let us consider $\kappa^x_{(A,n)}$. The corresponding possibilistic rule is (noticing $\alpha = 1 - 2^{-n}$)

$$\pi^x_{(A,\alpha)}(\omega) = 1 \text{ if } \omega \in A \text{ and } \pi(\omega) = \Pi(A)$$
$$= \pi(\omega) \text{ if } \omega \in A \text{ and } \pi(\omega) < \Pi(A)$$
$$= \pi(\omega) \text{ if } \omega \in \bar{A} \text{ and } \pi(\omega) < 1 - \alpha$$
$$= 1 - \alpha \text{ if } \omega \in \bar{A} \text{ and } \pi(\omega) \geq 1 - \alpha.$$

This definition can then be simplified

$$\pi^x_{(A,\alpha)}(\omega) = \pi(\omega \mid A) \text{ if } \omega \in A$$
$$= \min(1 - \alpha, \pi(\omega)) \text{ if } \omega \notin A.$$

This is the belief change operation (11) with $* = \min$: this operation has been given a clear meaning in Section 3.2 since $(A,n)$ is thus an unsure input that can be rejected if too uncertain.

The adjustment then can be expressed as follows in the possibilistic setting:

$$\pi^*_{(A,\alpha)} = \pi^-_A \text{ if } \alpha = 0$$
$$= (\pi^-_A)^x_{(A,n)} \quad \text{if } 0 < \alpha < N(A)$$
$$= \pi^x_{(A,n)} \quad \text{otherwise.}$$

Assume $N(A) > \alpha > 0$. Then:

$(\pi^-_A)^x_{(A,n)}(\omega) = \pi^-_A(\omega \mid A)$ if $\omega \in A$
$= \min(1 - \alpha, \pi^-_A(\omega))$ if $\omega \notin A$.

It is not difficult to check that

i) $\pi^-_A(\omega \mid A) = \pi(\omega \mid A)$. Indeed if $\omega \notin A$ then both sides are zero. If $\omega \in A$ then $\pi^-_A(\omega) = \pi(\omega)$ everywhere;

ii) if $\omega \notin A$ then $\max(\pi(\omega), \pi(\omega \mid \bar{A})) = \pi(\omega \mid \bar{A})$.

Hence $\pi^*_{(A,\alpha)}(\omega) = \pi(\omega \mid A)$ if $\omega \in A$
$= \min(1 - \alpha, \pi(\omega \mid \bar{A}))$ otherwise.

Assume now $N(A) \leq \alpha$; then, as stated above, $\pi^*_{(A,\alpha)} = \pi^x_{(A,\alpha)}$. But since $N(A) \leq \alpha$, $\Pi(A) \geq 1 - \alpha$, so when $\omega \notin A$

if $\pi(\omega) < 1 - \alpha$ then $\pi(\omega \mid \bar{A}) = \pi(\omega)$
if $\pi(\omega) = 1 - \alpha$ then $\pi(\omega \mid \bar{A}) \geq 1 - \alpha$

hence $\min(1 - \alpha, \pi(\omega)) = \min(1 - \alpha, \pi(\omega \mid \bar{A}))$ for $\omega \notin A$. It becomes clear that $\pi^*_{(A,\alpha)}$ takes the same form whether $N(A) > \alpha$ or not and coincides with the counterpart of Jeffrey's rule in the possibilistic setting, that is, rule (9) with $* = \min$. Note that it can be written in the form of a qualitative mixture (Dubois et al., 1993), very similar to Jeffrey's rule, when $\alpha > 0$

$$\pi^*_{(A,\alpha)}(\omega) = \max(\pi(\omega \mid A), \min(1 - \alpha, \pi(\omega \mid \bar{A}))).$$

This leads us to simplifying the expression of Williams adjustment into

$$\kappa^*_{(A,n)}(\omega) = \min(\kappa^*(\omega \mid A), \max(n, \kappa^*(\omega \mid \bar{A}))) \quad (12)$$

where $\kappa^*(\omega \mid A) = +\infty$ if $\omega \notin A$
$= \kappa(\omega)$ if $\kappa(\omega) > \kappa(A)$
$= 0$ if $\kappa(\omega) = \kappa(A)$

using the mappings $\pi(\omega) = 2^{-\kappa(\omega)}$ and $\alpha = 1 - 2^n$. This is equation (12) of the main text. The latter is the Spohnian version of the qualitative form of possibilistic conditioning. Turning $\kappa^*(\omega \mid A)$ into $\kappa(\omega \mid A) = \kappa(\omega) - \kappa(A)$ when $\omega \in A$ and max into sum, one gets the Spohnian $(A,n)$-conditioning rule. Note that when $n = 0$, $\kappa^*_{(A,n)}$ as per (12) does not recover the contraction $\kappa^-_A$, exactly for the same reason as its possibilistic counterpart (9). See the corresponding discussion. It seems somewhat artificial to enforce $\kappa^*_{(A,n)} = \kappa^-_A$ when $n = 0$, as done by Williams (1994b).